\documentclass[conference]{IEEEtran}
\IEEEoverridecommandlockouts

\usepackage{hyperref}
\hypersetup{breaklinks=true,colorlinks}  
\usepackage[symbol]{footmisc}       
\usepackage{microtype}              

\usepackage{bm}
\usepackage{amssymb}
\usepackage{amsmath}

\usepackage[binary-units=true,separate-uncertainty=true,multi-part-units=single]{siunitx}
\DeclareSIUnit\px{px}
\DeclareSIUnit\wattpeak{W_{P}}

\usepackage{subcaption}

\usepackage{cleveref}
\crefname{section}{Sec.}{Sections}
\crefname{figure}{Fig.}{Figure}
\crefname{table}{Tab.}{Table}
\crefname{equation}{Eq.}{Equation}
\crefname{appendix}{Appendix}{Appendix}

\usepackage{xspace}
\makeatletter
\DeclareRobustCommand\onedot{\futurelet\@let@token\@onedot}
\def\@onedot{\ifx\@let@token.\else.\null\fi\xspace}
\makeatother
\newcommand{\etal}[1]{#1~\textit{et~al\onedot}}

\usepackage{booktabs}

\usepackage{etoolbox} 
\usepackage{acronym}
\input{accaps.tex}

\newacro{PV}{photovoltaic}
\newacro{EL}{electroluminescense}
\newacro{IR}{infrared}
\newacro{IoU}{intersection over union}
\newacro{SVR}{support vector regression}
\newacro{DL}{deep-learning}
\newacro{MAE}{mean absolute error}
\newacro{MSE}{mean squared error}
\newacro{RMSE}{root mean squared error}
\newacro{MP}[PMPP]{power at maximum power point}
\newacro{MPP}{maximum power point}
\newacro{DNN}{deep neural network}
\newacro{SGD}{stochastic gradient descent}
\newacro{FC}{fully connected}
\newacro{CAM}{class activation map}
\newacro{GAP}{global average pooling}
\newacro{FCL}{fully connected layer}
\newacro{ZCA}{zero component analysis}
\newacro{t-SNE}{t-stochastic neighbour embedding}
\newacro{CV}{cross validation}
\newacro{STC}{standard test conditions}
\newacro{PL}{photoluminescence}

\usepackage[inline]{enumitem}
\setlist*[enumerate]{label=(\arabic*)}

\usepackage{csvsimple}

\newcommand{\myvector}[1]{\mathrm{#1}}
\newcommand{\mymatrix}[1]{\bm#1}

\newcommand{\typea}{T1\xspace} 
\newcommand{\typeb}{T2\xspace} 
\newcommand{\typec}{T3\xspace} 
\newcommand{\typed}{T4\xspace} 
\newcommand{\typee}{T5\xspace} 
\newcommand{\typef}{T6\xspace} 

\newcommand{\resneta}{ResNet18\xspace}

\newcommand{\imagenet}{\resneta~(I)\xspace} 
\newcommand{\pvpower}{\resneta~(I$\to$P)\xspace}  

\newcommand{\fcam}{\ensuremath{\myvector{{f}_{\text{map}}}}\xspace}

\newcommand{\pnom}{\ensuremath{P_{\text{nom}}}\xspace}
\newcommand{\prel}{\ensuremath{y}\xspace}
\newcommand{\prelest}{\ensuremath{\hat{y}}\xspace}

\newcommand{\mae}{\ensuremath{\text{MAE}}\xspace}
\newcommand{\rmse}{\ensuremath{\text{RMSE}}\xspace}

\newcommand{\pmpp}{\ensuremath{P_{\text{mpp}}}\xspace}
\newcommand{\pmppest}{\ensuremath{\hat{P}_{\text{mpp}}}\xspace}
\newcommand{\nsamples}{\ensuremath{N}\xspace}

\newcommand{\mywd}{\ensuremath{\lambda}\xspace}

\newcommand{\embedding}{\ensuremath{\myvector{f}_{\text{emb}}}\xspace}
\newcommand{\weights}{\ensuremath{\mymatrix{W}}\xspace}
\newcommand{\fclayer}{\ensuremath{\weights_{\text{fc}}}\xspace}
\newcommand{\sample}{\ensuremath{\myvector{x}}\xspace}

\newcommand{\resultsmaew}[1]{\csvloop{
                    file=data/051_plpower_summary/summary.csv,
                    head to column names,
                    filter strcmp={\model}{#1},
                    command={\SI[separate-uncertainty=true, multi-part-units=single]{\maeW(\uncertaintyW)}{\wattpeak}},
                }%
}
\newcommand{\resultsmae}[1]{\csvloop{
                    file=data/051_plpower_summary/summary.csv,
                    head to column names,
                    filter strcmp={\model}{#1},
                    command={\SI[separate-uncertainty=true, multi-part-units=single]{\mae(\uncertainty)}{\percent}},
                }%
}

\usepackage{tikz}
\usetikzlibrary{fit}
\usetikzlibrary{patterns}
\usetikzlibrary{calc}
\usetikzlibrary{positioning}
\usetikzlibrary{arrows.meta}
\usetikzlibrary{shapes}
\usetikzlibrary{spy}
\usepackage{rotating}
\usepackage{verbatim}
\usepackage{pgfplots}
\pgfplotsset{compat=newest}
\usepgfplotslibrary{statistics}
\usepackage{pgfplotstable}
\usepgfplotslibrary{external,patchplots}
\usepgfplotslibrary{colorbrewer,groupplots}

\pgfdeclarelayer{bg}
\pgfsetlayers{bg,main}  

\pgfplotsset{
    discard if/.style 2 args={
        x filter/.append code={
            \edef\tempa{\thisrow{#1}}
            \edef\tempb{#2}
            \ifx\tempa\tempb
                
            \fi
        }
    },
    discard if not/.style 2 args={
        x filter/.append code={
            \edef\tempa{\thisrow{#1}}
            \edef\tempb{#2}
            \ifx\tempa\tempb
            \else
                
            \fi
        }
    },
    discard boxplot if not/.style 2 args={
        /pgfplots/boxplot/data filter/.code={
            \edef\tempa{\thisrow{#1}}
            \edef\tempb{#2}
            \ifx\tempa\tempb
            \else
                
            \fi
        }
    }
}

\definecolor{colort1}{RGB}{228,26,28}
\definecolor{colort2}{RGB}{55,126,184}
\definecolor{colort3}{RGB}{77,175,74}
\definecolor{colort4}{RGB}{152,78,163}
\definecolor{colort5}{RGB}{255,127,0}
\definecolor{colort6}{RGB}{230,171,2}

\pgfplotsset{
    scatterclasses style/.style={
        scatter/classes={
            t1={mark=*,colort1,scale=1.5},
            t2={mark=*,colort2,scale=1.5},
            t3={mark=*,colort3,scale=1.5},
            t4={mark=*,colort4,scale=1.5},
            t5={mark=*,colort5,scale=1.5},
            t6={mark=*,colort6,scale=1.5}
        },
    },
}
\pgfplotsset{
    scatterplotnew style/.style={
        enlargelimits=false,
        axis on top,
        xlabel={\pmpp [\si{\wattpeak}]},
        ylabel={\pmppest [\si{\wattpeak}]},
        scatterclasses style,
        xmin=120,
        xmax=370,
        ymin=120,
        ymax=370,
    },
}
\newcommand{\scattercommon}{
    \addplot [
        domain=0:400,
        samples=2,
        no markers,
    ] {x};
    \addplot [
        domain=0:400,
        samples=2,
        no markers,
        dashed,
        white!50!black,
    ] {x+15};
    \addplot [
        domain=0:400,
        samples=2,
        no markers,
        dashed,
        white!50!black,
    ] {x-15};
}

\pgfplotsset{
    scatter samples/.style={
        only marks,
        scatter,
        scatter src=explicit symbolic,
        mark options={scale=0.5},
    }
}

\def\BibTeX{{\rm B\kern-.05em{\sc i\kern-.025em b}\kern-.08em
    T\kern-.1667em\lower.7ex\hbox{E}\kern-.125emX}}

\begin{document}

\title{Module-Power Prediction from PL Measurements using Deep Learning
\thanks{This project has been supported by IBC Solar AG and the German Federal Ministry for Economic Affairs and Energy (FKZ: 0324286 (iPV4.0) and 032429A (COSIMA)). Furthermore, we acknowledge the PV-Tera grant by the Bavarian State Government (No. 44-6521a/20/5).}
}

\renewcommand{\IEEEauthorrefmark}[1]{\textsuperscript{#1}}

\author{%
    \IEEEauthorblockN{%
        Mathis Hoffmann$^{1,2}$, 
        Johannes Hepp$^{2,3}$, 
        Bernd Doll$^{2,3,4}$,
        Claudia Buerhop-Lutz$^{3}$, 
        Ian Marius Peters$^{3}$,\\
        Christoph Brabec$^{2,3,4}$,
        Andreas Maier$^{1,4}$, 
        and Vincent Christlein$^{1}$
    }
    \IEEEauthorblockA{%
        \IEEEauthorrefmark{1}Pattern Recognition Lab, Friedrich-Alexander-Universität Erlangen-Nürnberg, Erlangen, 91058, Germany (FAU) \\
        \IEEEauthorrefmark{2}Materials for Electronics and Energy Technology, FAU \\
        \IEEEauthorrefmark{3}Helmholtz Institut Erlangen Nürnberg, Erlangen, 91058, Germany \\
        \IEEEauthorrefmark{4}Graduate School of Advanced Optical Technologies, Erlangen, 91058, Germany
    }%
}

\maketitle

\begin{abstract}
\sisetup{detect-weight=true, detect-family=true} 
The individual causes for power loss of photovoltaic modules are investigated for quite some time. Recently, it has been shown that the power loss of a module is, for example, related to the fraction of inactive areas. While these areas can be easily identified from \ac{EL} images, this is much harder for \ac{PL} images. With this work, we close the gap between power regression from \ac{EL} and \ac{PL} images. We apply a deep convolutional neural network to predict the module power from \ac{PL} images with a \ac{MAE} of \resultsmae{pvpower} or \resultsmaew{pvpower}. Furthermore, we depict that regression maps computed from the embeddings of the trained network can be used to compute the localized power loss. Finally, we show that these regression maps can be used to identify inactive regions in \ac{PL} images as well.
\end{abstract}


\acresetall

\section{Introduction}

\ifdefined\isabstract\else Recently, \ac{PV} power production has grown significantly as a result of the countermeasures to fight global warming. For example, the worldwide production has grown from \SI{190}{\tera\watt\hour} in 2014 to \SI{720}{\tera\watt\hour} in 2019~\cite{iea2019}. This is an increase by \SI{379}{\percent}.\fi To ensure constant performance of the power plants, regular inspection is required, since modules might be damaged during manufacturing, transport or installation. This raises the need for fast, accurate and non-invasive inspection methods.

In the last years, \ac{EL} has been widely adopted by the community as a useful tool to conduct inspection of solar modules~\ifdefined\isabstract\cite{kontges2011crack,mayr2019weakly}\else\cite{kontges2011crack,paggi2016global,mayr2019weakly,stromer2019enhanced,deitsch2019automatic}\fi. It allows to identify many types of defects. In particular, disconnected parts of the solar module that to not contribute to the power production (inactive areas), clearly stand out~\cite{buerhop2018evolution}. Previous works have shown that\ifdefined\isabstract\else~the number of cracks is loosely correlated to the power loss~\cite{dubey2018site} and that\fi~the power loss of a module is proportional to the fraction of inactive area, as long as it remains small~\cite{schneller2018electroluminescence}. Recently, \etal{Hoffmann}~\cite{hoffmann2020deep} used deep learning to determine the module power from \ac{EL} measurements. They introduce a visualization technique that allows to quantify the power loss of individual defects or cells as predicted by the model.

\begin{figure}[t]%
    \centering%
    \begin{subfigure}{0.49\linewidth}%
        \includegraphics[width=\linewidth,trim=1045 510 0 130,clip]{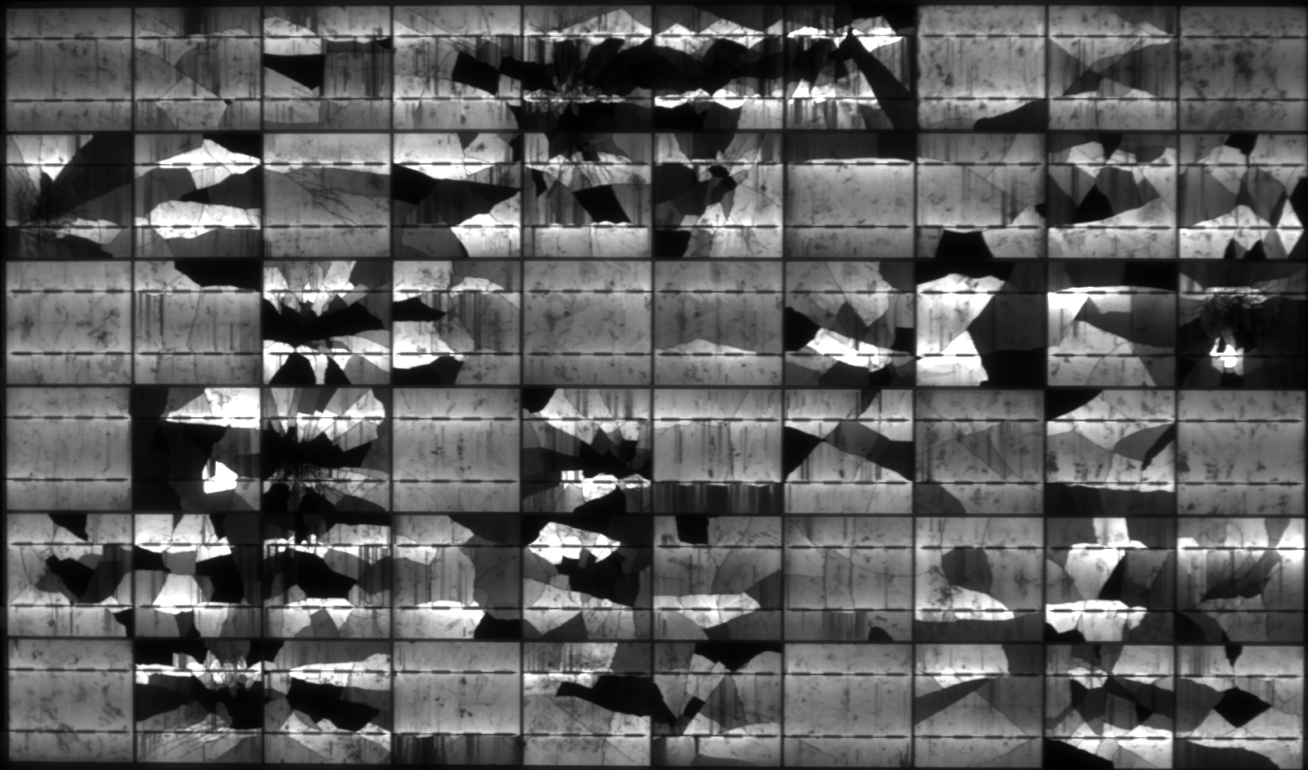}%
    \end{subfigure}%
    \hfill
    \begin{subfigure}{0.49\linewidth}%
        \includegraphics[width=\linewidth,trim=1045 510 0 130,clip]{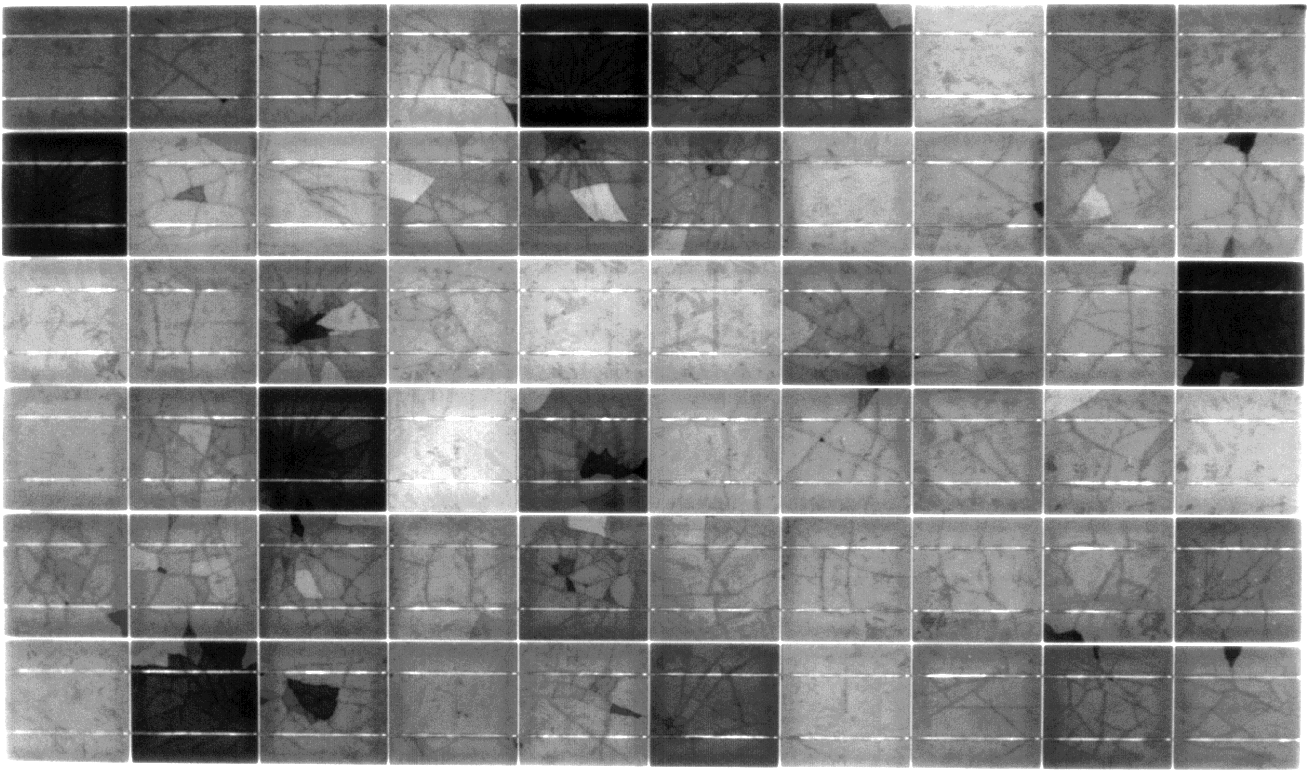}%
    \end{subfigure}%
    \caption{Comparison between \acl{EL} (left) and \acl{PL} (right) image of the same module.}%
    \label{fig:example}%
\end{figure}

However, \ac{EL} imaging comes at a price, since it requires to disconnect and power the string or module. Only recently, \ac{PL} imaging has become popular on an industrial scale. As opposed to \ac{EL}, the modules are excited by a light source and no external powering of modules is required. On the downside, inactive areas do not always show as black areas any more~(\cref{fig:example}). Instead, they appear with various different intensity levels. This has been previously reported by \etal{Doll}~\cite{doll2020contactless}.

In this work, we show that the deep learning-based approach~\cite{hoffmann2020deep} can be used to determine the power from \ac{PL} images of a module, too. To this end, we compile a dataset of \num{54} module \ac{PL} images along with measurements of the peak power $\pmpp$ and retrain the method using the new data. Furthermore, we investigate, if fine-tuning the models that have been released~\cite{hoffmann2020deep} can improve the performance even further. Finally, we show that the visualization technique using \acp{CAM} can be used to identify the inactive areas on \ac{PL} images.


\section{Methodology}\label{sec:methodology}

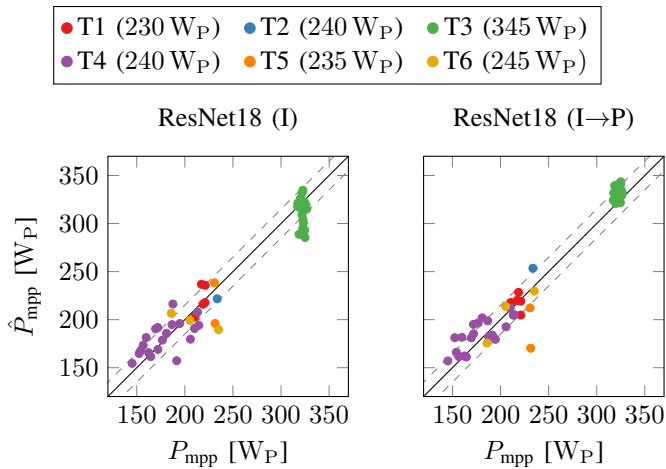
\begin{figure}[t]%
    \pgfplotsset{
        compat=newest,
        every axis/.append style={
            scatterplotnew style,
            xtick={150,200,250,300,350},
            ytick={150,200,250,300,350},
            title style={align=center, font=\linespread{0.8}\selectfont},
        }
    }%

    \centering%

    \begin{NoHyper}
    \ref{scatter-all-legend}
    \end{NoHyper}%
    \vspace{.2cm}%

    \begin{tikzpicture}
        \begin{groupplot}[
            group style={
                group size=2 by 1,
                vertical sep=1.5cm,
                ylabels at=edge left,
                xlabels at=edge bottom,
                yticklabels at=edge left,
                horizontal sep=1cm,
            },
            width=1/2*\linewidth + 0.35cm,
            height=1/2*\linewidth + 0.35cm,
        ]

            
            \nextgroupplot[
                title={\imagenet},
            ]
                \addplot [
                    scatter samples,
                ]table [x=peak_power, y=predicted_power, meta=source_key, col sep=comma]{data/051_plpower_summary/resnet18_init_imagenet.csv};
                \scattercommon
            
            \nextgroupplot[
                title={\pvpower},
                legend columns=3,
                legend to name=scatter-all-legend,
                legend entries={\typea~(\SI{230}{\wattpeak}),\typeb~(\SI{240}{\wattpeak}),\typec~(\SI{345}{\wattpeak}),\typed~(\SI{240}{\wattpeak}),\typee~(\SI{235}{\wattpeak}),\typef~(\SI{245}{\wattpeak)}},
                legend style={/tikz/every even column/.append style={column sep=0.25cm}},
            ]

                \addlegendimage{only marks,mark=*,colort1}; \label{pgfplots:mark-1}
                \addlegendimage{only marks,mark=*,colort2}; \label{pgfplots:mark-2}
                \addlegendimage{only marks,mark=*,colort3}; \label{pgfplots:mark-3}
                \addlegendimage{only marks,mark=*,colort4}; \label{pgfplots:mark-4}
                \addlegendimage{only marks,mark=*,colort5}; \label{pgfplots:mark-5}
                \addlegendimage{only marks,mark=*,colort6}; \label{pgfplots:mark-6}
                \addlegendimage{no marks,black}; \label{pgfplots:line-ideal}
                \addlegendimage{no marks,dashed,white!50!black}; \label{pgfplots:line-dashed}

                \addplot [
                    scatter samples,
                ]table [x=peak_power, y=predicted_power, meta=source_key, col sep=comma]{data/051_plpower_summary/resnet18_init_pvpower.csv};
                \scattercommon

        \end{groupplot}
    \end{tikzpicture}

    \caption{Distribution of estimation errors comparing pretraining on ImageNet~\cite{deng2009imagenet} or pretraining on the ELPVPower dataset~\cite{juelich2020pvpowerdata}. We aggregate the results from all three folds of the \acl{CV}. As a result, all \num{54} samples from the dataset are shown here. The module types are differentiated by color and the respective nominal powers are given by the legend. In addition, the ideal regression line~\ref{pgfplots:line-ideal} and the \SI[separate-uncertainty=true, multi-part-units=single]{0(15)}{\wattpeak} isoline~\ref{pgfplots:line-dashed} are shown.}
    \label{fig:scatter-all}
\end{figure}

We aim to estimate the power at the maximum power point \pmpp under STC conditions. Here, we use the same approach proposed by \etal{Hoffmann}~\cite{hoffmann2020deep} and estimate \pmpp relative to the nominal power \pnom. This is a sensible approach, since it assures that the estimates \prelest are in a similar scale, independent of the nominal power of a module. Then, we obtain the absolute power as
\begin{equation}
    \pmppest = \prelest\cdot\pnom\,.
\end{equation}
In this work, \prelest is computed by linear regression from the embedding of a \resneta~\cite{he2016identity}:
\begin{equation}
    \prelest = \fclayer\embedding\,,
\end{equation}
where $\embedding\in\mathbb{R}^{512}$ is given by $\embedding = f(\sample,\weights)$, $f$ represents the \resneta, \weights the parameters of $f$ and \fclayer is the linear regression.

We jointly optimize the parameters of the linear regression \fclayer and the parameters of the \resneta to minimize the mean squared error for all \nsamples samples in the training dataset, which is given by
\begin{equation}
    L = \frac{1}{\nsamples}\sum_{i=1}^\nsamples(\prelest_i-\prel_i)^2\,.
\end{equation}
As common with deep learning approaches, this is done by batch gradient descent. Except for the weight decay \mywd, we use the same hyperparameters that have been reported in the prior work. We found that the network seriously overfits to the training data with the reported setting for \mywd. This is explained by the dataset size, which is much smaller compared to the PVPower dataset~\cite{juelich2020pvpowerdata} used with the reference approach. We heuristically set $\mywd = 0.1$, which consistently gives good results.

\subsection*{Regression maps}\label{subsec:regression-maps}

\etal{Hoffmann}~\cite{hoffmann2020deep} propose to use a modified variant of \acp{CAM} to compute regression maps that give rise to a localized quantification of power losses. In the conventional \resneta, \embedding is computed by averaging over the \num{512} feature maps. This is commonly referred to as global average pooling. However, by averaging over the spatial dimensions of the feature maps, the spatial information, which is needed for the localized quantification of power losses, is lost. To this end, they propose to apply a $1\times1$ convolution to the feature maps, reducing the \num{512} maps to a single one, while preserving the spatial information.\ifdefined\isabstract~The resulting map is then used to compute \prelest during training and gives the localized power loss at test time.\else~Then, they compute the absolute value and multiply the result by $-1$. This ensures that the resulting regression map \fcam is strictly negative. The relative power \prelest is then computed as
\begin{equation}
    \prelest = 1 + \sum_{i,j \in \Omega_{\myvector{f}}} \underbrace{-\text{ReLU}(\myvector{f}_{i,j})}_{\fcam}\,.
\end{equation}
This way, the network can be trained using only the relative power of the module as supervision signal, while the localized power loss is obtained in a weakly supervised manner. The power loss per cell is computed by integrating over the corresponding area of \fcam. This approach has shown promising results on \ac{EL} images already. In the following, we show that it performs well on \ac{PL} images as well.\fi

\section{Experiments}\label{sec:experiments}

\ifdefined\isabstract

\begin{figure*}[tp]
    \centering
    \begin{tikzpicture}
        \begin{groupplot}[
            group style={
                group size=2 by 1,
                vertical sep=1.5cm,
                horizontal sep=0.25cm,
                yticklabels at=edge left,
            },
            width=1/1.77*\textwidth,
            height=1/3.4*\textwidth,
        ]
        
        \nextgroupplot[
            xtick={64,192,320,448,576},
            xticklabels={1,2,3,4,5},
            ytick={320,448,576,708},
            yticklabels={A,B,C,D,E},
            axis x line*=top,
            y dir=reverse,
            colormap/viridis,
            point meta min=0,
            point meta max=100,
            colorbar horizontal,
            colorbar style={
                xtick={0, 25, 50, 75, 100},
                width=0.5*\pgfkeysvalueof{/pgfplots/parent axis width},
                xlabel=Relative power~[\%],
            },
            enlargelimits=false,
            xmin=0,
            xmax=654,
            ymin=384,
            ymax=640
        ]
            \addplot graphics[xmin=0, ymin=384, xmax=654, ymax=640,
                includegraphics={trim=0 128 654 384, clip}] {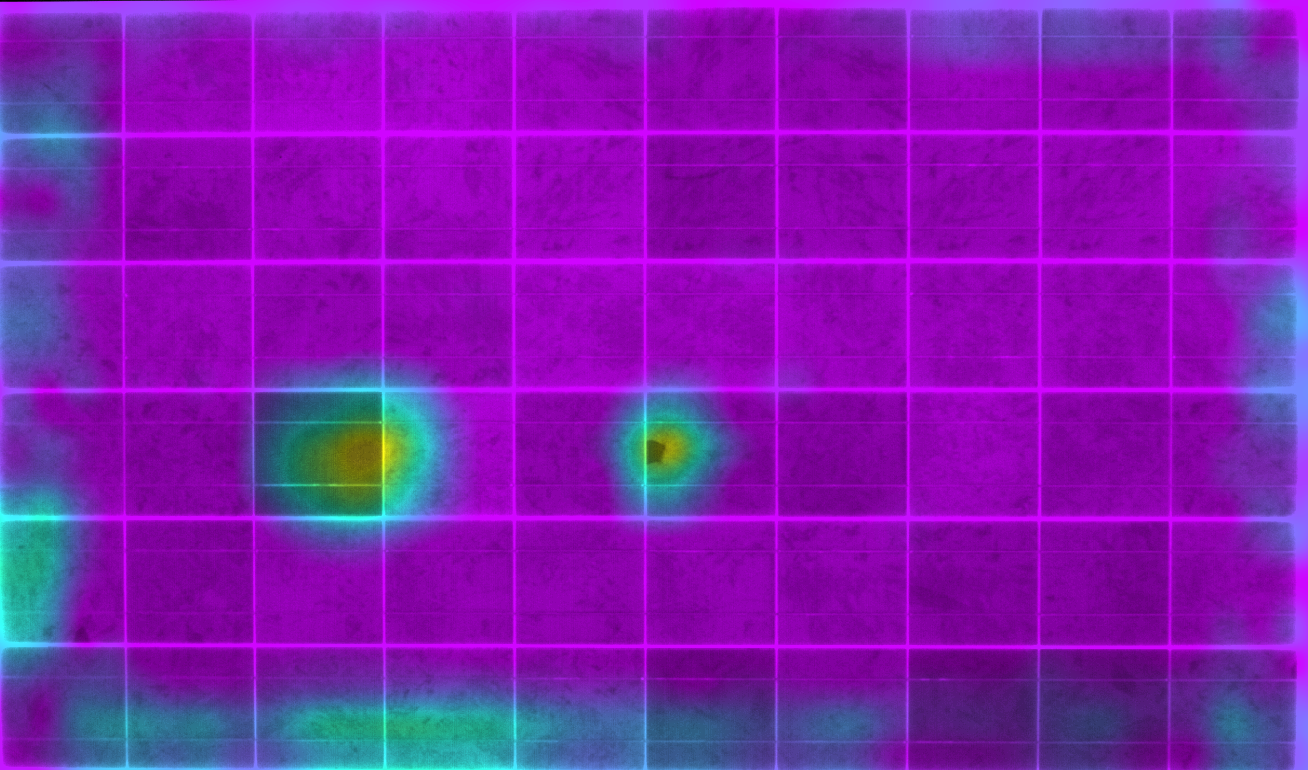};

            \addplot [
                mark=none,
                draw=none,
                nodes near coords={\small\num[round-mode=places, round-precision=1, tight-spacing=true]{\pgfplotspointmeta}},
                nodes near coords align={anchor=center},
                nodes near coords style={
                    rectangle,
                    fill=white,
                    fill opacity=0.5,
                    text opacity=1,
                    inner sep=3pt,
                    rounded corners=2pt,
                },
            ] 
            table[
                x=x,
                y=y,
                meta=v,
                col sep=comma,
                point meta=explicit symbolic
                ] {data/052_finetune_cam/32.csv};

        \nextgroupplot[
            xtick={64,192,320,448,576},
            xticklabels={1,2,3,4,5},
            axis x line*=top,
            ytick style={draw=none},
            y dir=reverse,
            colormap/blackwhite,
            point meta min=0,
            point meta max=28000,
            colorbar horizontal,
            colorbar style={
                xtick={0, 7000, 14000, 21000, 28000},
                scaled ticks=false,
                xlabel=Intensity~[Counts],
                width=0.5*\pgfkeysvalueof{/pgfplots/parent axis width},
            },
            enlargelimits=false,
            xmin=0,
            xmax=654,
            ymin=384,
            ymax=640
        ]
            \addplot graphics[xmin=0, ymin=384, xmax=654, ymax=640,
                includegraphics={trim=0 128 654 384, clip}] {data/EL/032_0_28000_resized.png};

        \end{groupplot}
    \end{tikzpicture}

    \caption{Quantification of the per-cell power loss using \acp{CAM} from a modified \resneta (left) in comparison to an \ac{EL} image of the same module (right). We color-code the original \ac{EL} measurement with the given colormap. For every cell, we report the power loss determined by the model in \si{\wattpeak}.}
    \label{fig:cam}
\end{figure*}

\else

\begin{figure*}[tp]
    \centering
    \begin{tikzpicture}
        \begin{groupplot}[
            group style={
                group size=2 by 1,
                vertical sep=1.5cm,
                horizontal sep=0.25cm,
                yticklabels at=edge left,
            },
            width=1/1.77*\textwidth,
            height=1/2*\textwidth,
        ]
        
        \nextgroupplot[
            xtick={64,192,320,448,576},
            xticklabels={1,2,3,4,5},
            ytick={320,448,576,708},
            yticklabels={A,B,C,D,E},
            axis x line*=top,
            ytick style={draw=none},
            y dir=reverse,
            colormap/blackwhite,
            point meta min=0,
            point meta max=28000,
            colorbar horizontal,
            colorbar style={
                xtick={0, 7000, 14000, 21000, 28000},
                xlabel=Intensity~[Counts],
                width=0.5*\pgfkeysvalueof{/pgfplots/parent axis width},
            },
            enlargelimits=false,
            xmin=0,
            xmax=654,
            ymin=256,
        ]
            \addplot graphics[xmin=0, ymin=256, xmax=654, ymax=770,
                includegraphics={trim=0 0 654 256, clip}] {data/EL/032_0_28000_resized.png};

        \nextgroupplot[
            xtick={64,192,320,448,576},
            xticklabels={1,2,3,4,5},
            ytick={},
            axis x line*=top,
            y dir=reverse,
            colormap/viridis,
            point meta min=0,
            point meta max=100,
            colorbar horizontal,
            colorbar style={
                xtick={0, 25, 50, 75, 100},
                width=0.5*\pgfkeysvalueof{/pgfplots/parent axis width},
                xlabel=Relative power~[\%],
            },
            enlargelimits=false,
            xmin=0,
            xmax=654,
            ymin=256,
        ]
            \addplot graphics[xmin=0, ymin=256, xmax=654, ymax=770,
                includegraphics={trim=0 0 654 256, clip}] {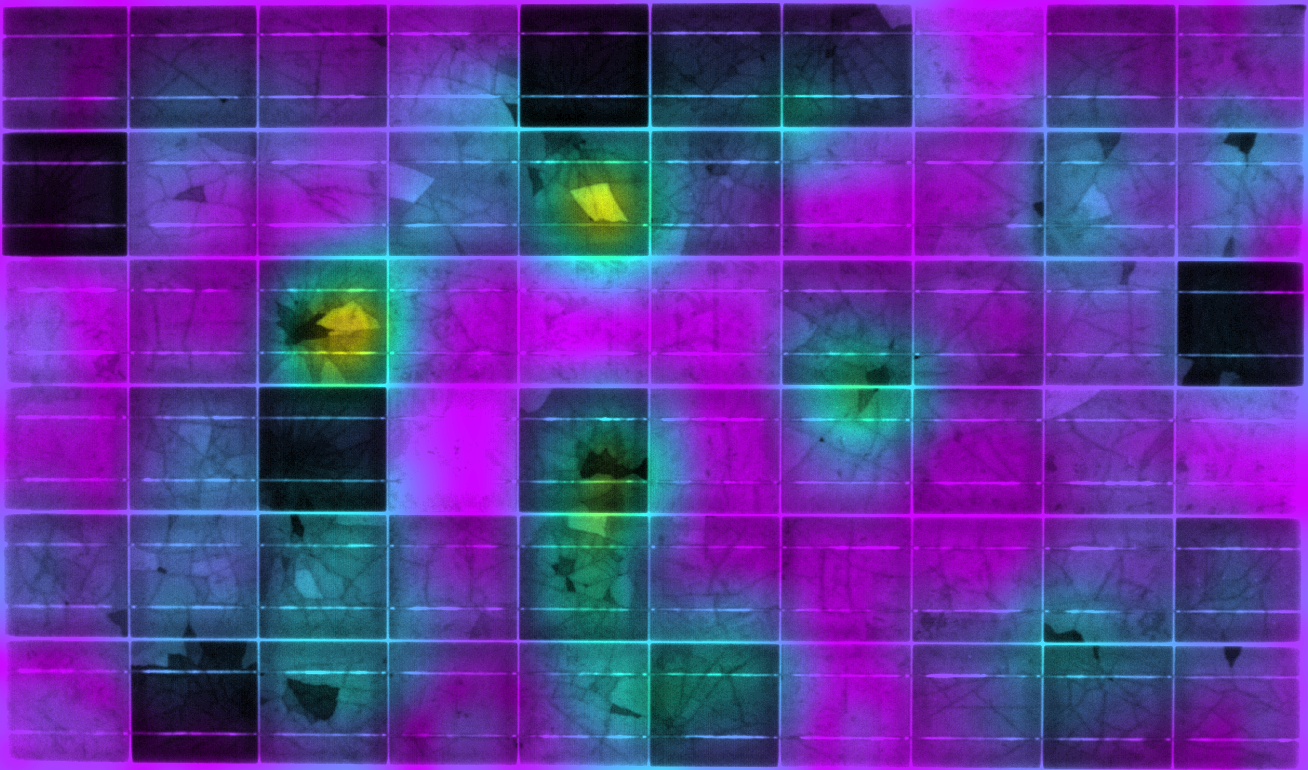};

            \addplot [
                mark=none,
                draw=none,
                nodes near coords={\small\num[round-mode=places, round-precision=1, tight-spacing=true]{\pgfplotspointmeta}},
                nodes near coords align={anchor=center},
                nodes near coords style={
                    rectangle,
                    fill=white,
                    fill opacity=0.5,
                    text opacity=1,
                    inner sep=3pt,
                    rounded corners=2pt,
                },
            ] 
            table[
                x=x,
                y=y,
                meta=v,
                col sep=comma,
                point meta=explicit symbolic
                ] {data/052_finetune_cam/32.csv};

        \end{groupplot}
    \end{tikzpicture}

    \caption{Quantification of the per-cell power loss using \acp{CAM} from a modified \resneta (right) in comparison to an ac{EL} image of the same module (left). We color-code the original \ac{EL} measurement with the given colormap. Note that brighter colors correspond to regions with high relative power loss. Color-coding is done such that the intensity is given by the original image. As a result, color appearence does not exactly correspond to the legend. For every cell, we integrate over the corresponding area of the \ac{CAM} and multiply the result by \pnom. This results in the power loss determined by the model in \si{\wattpeak}.}
    \label{fig:cam}
\end{figure*}

\fi

In our experiments, we focus on two aspects. First, we show that the power of a module can be estimated from a \ac{PL} image of a module despite the fact that inactive areas are not clearly visible from those images in any case. Second, we show that the regression maps can be used to locate disconnected areas on the module.

\subsection{Data}

For our experiments, we use a small dataset of only \num{54} \ac{PL} images. These images have been recorded under lab conditions with a front-illuminated SI camera with $2048^2$ pixels. Photo excitation has been conducted using our LED PL setup, which has been previously reported~\cite{doll2020contactless}.

The dataset covers \num{6} different types of modules, which we denote as \typea-\typef, with nominal powers ranging from \SIrange[]{230}{345}{\wattpeak} and maximum powers ranging from \SIrange[]{145}{327}{\wattpeak}. The modules of type \typea-\typeb and \typed-\typef feature \num{60} cells arranged in \num{10} columns, while \typec has \num{72} cells in \num{12} columns.

\ifdefined\isabstract
\else
Prior to processing by the network, images are preprocessed. Here, images are cropped and scaled to a common resolution. Furthermore, they are normalized such that the mean intensity $\mu$ over all images computes as $\mu=0$ and the standard deviation $\sigma$ is $\sigma=1$. During training, we apply online data augmentation similar to the reference method. This includes random horizontal and vertical flips as well as slight rotations of the images.
\fi

\subsection{Results}

\begin{table}[t]
    \centering
    \caption{Results computed by \acl{CV}. The numbers are averaged over all test folds. Furthermore, we show the standard deviation of the MAE.}
    \csvloop{
        file=data/051_plpower_summary/summary.csv,
        head to column names=false,
        column names={model=\model,maeW=\maeW,mae=\mae,stdW=\stdW,std=\std,rmseW=\rmseW,rmse=\rmse,texmodel=\texmodel,uncertainty=\uncertainty,uncertaintyW=\uncertaintyW},
        before reading=\sisetup{table-format=2.1,round-mode=places,round-precision=1,table-number-alignment=left},
        tabular={lS[table-format=1.1,separate-uncertainty=true,table-figures-uncertainty=3,table-align-uncertainty=true]S[table-format=2.1,separate-uncertainty=true,table-figures-uncertainty=3,table-align-uncertainty=true]S[table-format=2.1,table-number-alignment=left]S[table-format=2.1,table-number-alignment=left]},
        table head={\toprule \linespread{0.1} & \multicolumn{2}{c}{MAE} & \multicolumn{2}{c}{RMSE} \\ \linespread{0.1} & [\si{\percent}] & [\si{\wattpeak}] & [\si{\percent}] & [\si{\wattpeak}] \\ \midrule},
        command=\texmodel & \mae(\uncertainty) & \maeW(\uncertaintyW) & \rmse & \rmseW,
        table foot=\bottomrule
    }
    \label{tab:results}
\end{table}

Since the dataset is small, it is challenging to draw meaningful conclusions from the result. To overcome this issue, we conduct a three-fold \ac{CV} and join the results of all folds. Here, we perform a stratified split, such that the distribution of \prel is similar for all three folds.

We train two different variants of the model. First, we stick to the procedure from the reference method and initialize the network with weights computed by pretraining on ImageNet. We denote this variant \imagenet. Second, we use the weights that has been published with the reference implementation for initialization. Since this has first been trained on ImageNet and then finetuned on the PVPower dataset~\cite{juelich2020pvpowerdata}, we denote this variant \pvpower. The results are summarized by~\cref{tab:results} and~\cref{fig:scatter-all}. Here, we also include a baseline that is computed by calculating the mean of \prel over every sample of the respective training set and use the result as the prediction \prelest for every sample of the corresponding test set. This gives a lower bound to the error.\ifdefined\isabstract\else~Every model that is better than weighted random predictions should surpass this lower bound.\fi~From~\cref{tab:results} we see that, despite the very small dataset, both variants perform much better than the baseline. Furthermore, we observe that pretraining on the PVPower dataset improves the results slightly.

Finally, we show and exemplary regression map in~\cref{fig:cam} and compare it to the \ac{EL} image of the same module. In summary, we see that the magnitude of predicted power loss per cell is consistent to the amount of inactive area as seen from the \ac{EL} image, although the inactive area is not always visible in the \ac{PL} image. For example, cells C1 and C2 have a similar appearance in the \ac{PL} image, although C2 is damaged more severely, which can only be seen from the \ac{EL} image.\ifdefined\isabstract\else~However, the model prediction is consistent to the \ac{EL} image, since C2 is predicted to have a higher power loss.\fi~Furthermore, we find that the model recognizes that inactive areas might appear as darker or brighter regions. This can be seen from cells B3 and C5. \ifdefined\isabstract\else~Although B3 is mostly dark in the \ac{PL}, whereas C5 has only few dark spots, they are predicted to have a similar power loss.\fi

\ifdefined\isabstract
\section{Discussion of the Significance of this Work for the Field}

The determination of the power of a module is of significant interest, since it is vital to the efficient operation of \ac{PV} power plants. Only recently, \ac{EL} imaging has been used to determine the power of a module~\cite{hoffmann2020deep}. However, this requires to disconnect every single module or string. \ac{PL} imaging provides efficient means to speed up measurements, since an external light source is used for excitation and no disconnection of modules is required any more. On the downside, \ac{PL} images complicate the interpretation, since inactive areas are not easily visible any more~\cite{doll2020contactless}. We show that power prediction can be performed using \ac{PL} imaging despite the challenging interpretability. Furthermore, we depict that deep learning is capable of learning relevant features using a relatively small dataset. Finally, we show that these features can be used to identify inactive areas as well.
\fi

\section{Summary}

We experimentally show that \ac{PL} images of solar modules can be used to determine the power of a module with a MAE of \resultsmaew{pvpower}, although inactive areas are not well represented by this modality. To this end, we compile a dataset of \num{54} \ac{PL} images along with their powers and train a deep neural network to predict the module power. Furthermore, we apply the approach by \etal{Hoffmann}~\cite{hoffmann2020deep} to compute regression maps that allow to quantify the localized power loss. Using these maps, we qualitatively show that the network learns the weakly supervised localization of inactive areas and that the results are consistent to reference \ac{EL} images.

We are confident that the quantitative results will become better, if a larger training dataset is used. Further, we believe that these preliminary results will amplify research in the field of \ac{PL} imaging for solar module inspection. For example, they can help to perform root cause analysis for damaged modules using \ac{PL} images only.

\bibliography{mybibfile}
\bibliographystyle{ieeetr}

\end{document}